# Applying Discrete PCA in Data Analysis

**Wray Buntine**
Complex Systems Computation Group,
Helsinki Institute for Information Technology
P.O. Box 9800, FIN-02015 HUT, Finland.
`Wray.Buntine@HIIT.FI`

**Aleks Jakulin**
Faculty of Computer and Information Science
University of Ljubljana
Tržaška 25, SI-1001, Ljubljana, Slovenia
`Jakulin@IEEE.ORG`

## Abstract

Methods for analysis of principal components in discrete data have existed for some time under various names such as grade of membership modelling, probabilistic latent semantic analysis, and genotype inference with admixture. In this paper we explore a number of extensions to the common theory, and present some application of these methods to some common statistical tasks. We show that these methods can be interpreted as a discrete version of ICA. We develop a hierarchical version yielding components at different levels of detail, and additional techniques for Gibbs sampling. We compare the algorithms on a text prediction task using support vector machines, and to information retrieval.

## 1 INTRODUCTION

Principal component analysis (PCA) latent semantic indexing, and independent component analysis (ICA) (Hyvärinen et al., 2001) are key methods in the statistical engineering toolbox. They have a long history, are used in many different ways, and under different names. They were primarily developed in the engineering community where the notion of a filter is common, and maximum likelihood methods less so. They are usually applied to measurements and real valued data.

Relatively recently the learning community has become aware of a seemingly similar approach for discrete data that appears under many names: grade of membership (Woodbury & Manton, 1982) used for instance in the social sciences, demographics and medical informatics, genotype inference using admixtures (Pritchard et al., 2000), probabilistic latent semantic indexing (Hofmann, 1999) latent Dirichlet allocation (Blei et al., 2003), and multiple aspect mod-elling for document analysis (Minka & Lafferty, 2002). These methods are equivalent, ignoring statistical methodology and notation. Note the representation of (Pritchard et al., 2000) is completely different, thus one also needs to translate notations. We refer to these methods jointly as *discrete PCA*.

In this paper we explore these methods in some common modes for data analysis, and explore some of the uses and relationships. We present a number of useful extensions that illustrate the versatility of these methods. Our contribution is to provide some additional techniques and methodology to allow wider use of these tools. For instance, with a small change, the methods turn to be a discrete version of ICA. Hierarchical components can also be built. The first author has developed software in C and coded for POSIX Linux that was used in these experiments and is available under the GNU GPL license from him.

## 2 THE BASIC MODEL

A good introduction to these models from a number of viewpoints is (Blei et al., 2003; Buntine, 2002). They are directly analogous to the Gaussian model of principal component analysis (Buntine, 2002). The simplest version consists of a linear admixture of different multinomials, and can be thought of as sampling words to make up a bag, as a representation for a document. The notation of words, bags and documents will be used throughout, even though other kinds of data also apply. In standard mixture models, each document in a collection is assigned a (hidden) topic. In this new model, each word in each document is assigned a (hidden) topic.

- We have a total count $L$ of words to sample.

- We partition these $L$ words into $K$ topics, components or aspects: $c_1, c_2, ...c_K$ where $\sum_{k=1,...,K} c_k = L$. This is done using a hidden proportion vector $\boldsymbol{m} = (m_1, m_2, ..., m_K)$. The in-tention is that, for instance, a sporting article may have 50 general vocabulary words, 40 words relevant to Germany, 50 relevant to football, and 30 relevant to people's opinions. Thus L=170 are in the document and the topic partition is (50,40,50,30).

- In each partition, we then sample words according to the multinomial for the topic, component or aspect. This is the base model for each component. This then yields a bag of word counts for the $k$-th partition, $\boldsymbol{w}_{k,\cdot} = (w_{k,1}, w_{k,2}, ..., w_{k,J})$. Here $J$ is the dictionary size, the size of the basic multinomials on words. Thus the 50 football words are now sampled into actual dictionary entries, "forward", "kicked", "covered" etc.

- The partitions are then combined additively, hence the term *admixture*, to make a distinction with classical mixture models. This yields the final sample of words $\boldsymbol{r} = (r_1, r_2, ..., r_J)$ by totalling the corresponding counts in each partition, $r_j = \sum_{k=1,...,K} w_{k,j}$. Thus if an instance of "forward" is sampled twice, as a football word and a general vocabulary word, then we return the count of 2 and its actual topical assignments are lost, they are hidden data.

The full probability model is then

$$
\begin{aligned}
\boldsymbol{m} &\sim \text{Dirichlet}(\boldsymbol{\alpha}) , \\
\boldsymbol{c} &\sim \text{Multinomial}(\boldsymbol{m}, L) , \\
\boldsymbol{w}_{k,\cdot} &\sim \text{Multinomial}(\boldsymbol{\Omega}_{k,\cdot}, c_k) \quad \text{for } k = 1, \ldots, K .
\end{aligned}
$$

The hidden or latent variables here are $\boldsymbol{m}$ and $\boldsymbol{w}$ for each document, whereas $\boldsymbol{c}$ is derived. The full likelihood for a single document $p(\boldsymbol{m}, \boldsymbol{w} \,|\, \boldsymbol{\alpha}, \boldsymbol{\Omega})$ then simplifies to:

$$
\frac{1}{Z_D(\boldsymbol{\alpha})} \prod_k m_k^{\alpha_k - 1} \prod_{k,j} m_k^{w_{k,j}} \Omega_{k,j}^{w_{k,j}} . \tag{1}
$$

A combinatoric term is also present or not, depending on whether data ordering is recorded or not (Buntine, 2002). The proportions $\boldsymbol{m}$ correspond to the hidden components for a document. The partitioned counts $\boldsymbol{w}$ correspond to the topic assignments given to individual words. Either $\boldsymbol{w}$ or $\boldsymbol{m}$ can be marginalized using standard distributional methods to yield another likelihood, however only the one of Equation (1) is a simple product of exponential family distributions.

## 2.1 ALGORITHMS

Neither of the three likelihoods yield to standard EM analysis. For instance, in Equation (1) the hidden variables $\boldsymbol{m}$ and $\boldsymbol{w}$ are coupled. Algorithms for this problem follow some of the usual approaches in the community, albeit with considerable sophistication:

- Annealed maximum likelihood (Hofmann, 1999), best viewed in terms of its clustering precursor (Hofmann & Buhmann, 1997),

- Gibbs sampling on $\boldsymbol{w}$, $\boldsymbol{m}$ and $\boldsymbol{\Omega}$ in turn using a full probability distribution (Pritchard et al., 2000),

- mean field methods (Blei et al., 2003), and

- expectation propagation (EP, like so-called cavity methods) (Minka & Lafferty, 2002).

In our experience, the mean field approach is fast but yields badly biased estimates of $\boldsymbol{m}$, and sometimes poor estimates of $\boldsymbol{\Omega}$. Note its speed and the likelihood of results will be considerably worse if repeated updates of $\boldsymbol{m}$ are done in initial cycles for each update of $\boldsymbol{\Omega}$, as some published implementation do. The Gibbs method is surprisingly fast but still a factor of 2-8 times slower than mean field; it requires a sample recording stage after burn-in, and it is difficult to compare convergence criteria. Gibbs is somewhat rapidly mixing because of the nature of the dual variables (the $\boldsymbol{\Omega}$ matrix is completely re-estimated from the counts $\boldsymbol{w}$ in each cycle, and $\boldsymbol{w}$ from the components $\boldsymbol{m}$). EP has unrealistic intermediate storage requirements, storing data per $w_{k,j}$, not needed by the other methods. For $K$ components and a collection with $T$ total words, storage required is $O(KT)$ which approaches terabytes on a standard sub-gigabyte collection. It also appears to be troubled by the Poisson regime that dominates in sparse word count data.

In experiments below we use Gibbs sampling. An initial burn-in period is done with $O(100)$ cycles. Thereafter, the $\boldsymbol{\Omega}$ parameters are averaged in the following $O(50)$ cycles. Note for this $\text{Exp}_{\boldsymbol{\Omega} \sim p(\boldsymbol{\Omega} | \boldsymbol{w}, \boldsymbol{m})}(\boldsymbol{\Omega})$ is averaged, not the raw $\boldsymbol{\Omega}$, for an unbiased but improved estimate. Then if document component data is wanted, the estimate of $\boldsymbol{\Omega}$ is fixed, and a further $O(50)$ cycles are done to estimate $\boldsymbol{m}$ or $\boldsymbol{w}$. When Gibbs for estimating $\boldsymbol{m}$ is done in one stage for each document, it is more efficient and more sophisticated sample information can be recorded about $\boldsymbol{m}$.

## 3 EXTENSIONS

In this paper we first point out several extensions to this basic form that make the method an attractive model family, we believe underutilized. We note, however, that one key aspect of the theory is underdeveloped: identifiability in the statistical sense.

## 3.1　MULTIVARIATE DATA

Discrete PCA extends easily to a set of separate multinomials: For instance, we could partition the words of a document up into five parts: title words, nouns in the body, verbs in the body, adjectives in the body, adverbs in the body, and ignore general function words and pronouns. Each document is then represented as five bags of words, and an admixture is made of the 5 multinomials separately, but according to the common component proportions $\boldsymbol{m}$.

This multinomial usage occurs in most earlier versions of the method, and is why the method is ideal for genotype data, demographics, voting records, medical informatics, etc. In the theory, the third multinomial on the $\boldsymbol{w}$ in the theory of Section 2 is repeated for the separate bags for a document. In implementation this changes one normalization step on counts, thus is simple to include.

## 3.2　DISCRETE INDEPENDENT COMPONENT ANALYSIS

With a small change, discrete PCA computes independent components. The full probability model is modified only in the generation of the count partition $\boldsymbol{c}$:

$$
\begin{aligned}
\lambda_k &\sim \text{Gamma}(\alpha_k, 1) \quad \text{for } k = 1, \dots, K , \\
c_k &\sim \text{Poisson}(\lambda_k) \quad \text{for } k = 1, \dots, K ,
\end{aligned}
$$

The full likelihood for a single document $p(\boldsymbol{\lambda}, \boldsymbol{w} \mid \boldsymbol{\alpha}, \boldsymbol{\Omega})$ then simplifies to:

$$
\prod_k \frac{1}{\Gamma(\alpha_k)} \lambda_k^{\alpha_k - 1} e^{-\lambda_k - 1} \prod_{k,j} \frac{1}{w_{k,j}!} \lambda_k^{w_{k,j}} \Omega_{k,j}^{w_{k,j}} . \quad (2)
$$

The components for a document have now changed from being a probability vector $\boldsymbol{m}$ to being $K$ independent Gamma variables $\lambda_k$. This is different from ICA however, in that these independent components are not a linear invertible transformation of the data but a lossy summary. In practice this is irrelevant. The engineering practice of ICA commonly uses derived data from a PCA of some original data, i.e., itself a summary. For instance, when applied to documents, word counts are first turned into TF-IDF scores, then summarized using PCA, and then ICA is performed. There is currently no version for discrete data for ICA[1].

Independent Gamma variables can always be normal-

ized to form a Dirichlet. This model is equivalent to:

$$
\begin{aligned}
\lambda &\sim \text{Gamma}\left(\sum_k \alpha_k, 1\right) , \\
L &\sim \text{Poisson}(\lambda) , \\
\boldsymbol{m} &\sim \text{Dirichlet}(\boldsymbol{\alpha}) , \\
\boldsymbol{c} &\sim \text{Multinomial}(\boldsymbol{m}, L) ,
\end{aligned}
$$

Thus we see that discrete PCA, by also sampling the total count $L$, can be interpreted as a discrete version of ICA. In implementation this requires some independent processing of the total counts but is otherwise identical. The resulting counts can also be rounded to produce nice data for indexing in a standard information retrieval engine.

## 3.3　ESTIMATING THE "RIGHT" NUMBER OF COMPONENTS $K$

We have developed a simple scheme using importance sampling in Gibbs to estimate the evidence term for a model. If the document data for the $i$-th document is $\boldsymbol{r}_i$, and there are $I$ documents, this evidence term is defined as $p(\boldsymbol{r}_1, \dots, \boldsymbol{r}_I \mid \text{discrete PCA}, K)$ and is traditionally the hardest part of Bayesian inference (note that Gibbs sampling is implicitly Bayesian). We would like to find the value of $K$ with the highest evidence. In popular terms, this could be used to find the "right" number of components, though in practice such a thing may not exist.

Pritchard *et al.* previously proposed an *ad hoc* approach to address this problem, and we need to note some practicalities about Gibbs sampling in such models (Pritchard et al., 2000). Because the model is symmetric in the $K$ components, and no symmetry breaking is done, in principle the results should be uniform as sampling cycles through all $K!$ ways in which one model could appear. Thus if Gibbs sampling worked perfectly in this application, the results would be useless without symmetry breaking. In practice, however, we converge to a single local optimum and Gibbs sampling explores the vicinity of this region, performing local averaging.

First, consider the theory of importance sampling applied to estimating the evidence for a model with parameters and hidden data $\theta$, and observed data $\boldsymbol{x}$. We wish to estimate $\int_\theta p(\boldsymbol{x}|\theta) p(\theta) \mathrm{d}\,\theta$. Using some sampling method we generate samples $\theta_1, \theta_2, \dots, \theta_N$ according to the posterior distribution $q(\theta) \propto p(\boldsymbol{x}|\theta) p(\theta)$. In importance sampling, the expected value of $u()$ is estimated using

$$
\text{Exp}_{\theta \sim p(\theta)}(u(\theta)) = \frac{\sum_n u(\theta_n) p(\theta_n) / q(\theta_n)}{\sum_n p(\theta_n) / q(\theta_n)}
$$

---

[1]Personal communication from A. Hyvärinen.

The lower normalizing term is used because we only have the proportionality for $q()$, not the normalizer. Applying this to evidence estimation, one gets the approximation

$$p(\boldsymbol{x}) \approx \frac{N}{\sum_n 1/p(\boldsymbol{x}|\theta_n)} .$$

Moreover, variational arguments show this is the importance sampler for evidence that minimizes estimation variance. Better sampled estimates for evidence cannot be found! Now Gibbs can be used instead here since it too computes averages. In our case, adding in the symmetric factor, this becomes

$$p(\boldsymbol{r}_1, \ldots, \boldsymbol{r}_I \,|\, \text{discrete PCA}, K) \approx$$
$$\frac{1}{K!} \frac{N}{\sum_n 1/p(\boldsymbol{r}_1, \ldots, \boldsymbol{r}_I | \boldsymbol{\Omega}_n, K)}$$

which is a byproduct of the sampling anyway. Thus, in as far as we know that Gibbs sampling is not properly mixing (or by symmetry our results would be useless anyway) and thus is just computing a local estimate, this is also computing a local estimate of evidence.

### 3.4   HIERARCHICAL COMPONENTS

The hierarchies or dendrograms generated by standard agglomerative clustering algorithms are frequently printed in applied science journals. It should be clear from the independence properties in Section 3.2 that the components in discrete PCA by design strongly resist any form of hierarchical ordering. Moreover, because the dimension of the underlying component multinomial $J$ (or dictionary size) can be comparable to the sample size, hierarchical sampling of the multinomial parameters $\boldsymbol{\Omega}_{k,\cdot}$ is statistically impractical. But hierarchical ordering is important if a human wishes to interpret results by inspection the typical content of each component.

Hierarchies in discrete PCA can be designed in by forcing a hierarchical correlation on the proportion vector $\boldsymbol{m}$. A simple example is as follows: We have a root component. All documents share a sizeable proportion of this, perhaps 5-10% rather than 100/K% proportion each component usually gets on average. The root component then record statistics about stop words and such. In newspaper content this records things like days of the week, quoting verbs like "said", etc. The components below it them divvy up the remaining content. The general hierarchy only forces one kind of constraint on the proportions $\boldsymbol{m}$: if a child's proportion is large, then its parent's proportion is also going to be large. Children are constrained to correlate with their parents. Thus if the parent node were "general sports" and the children "soccer," "cricket,"

"baseball," etc., then the occurrence of a large number of soccer words in a document means, with high probability, there will be a higher than normal proportion of general sports words in the document as well. This is done by enforcing correlations between $\boldsymbol{m}$ according to their position in the hierarchy.

The basic model then is changed only in the way $\boldsymbol{m}$ is sampled. Represent the components $1, \ldots, K$ as a tree, where each index $k$ has its parents, children, ancestors, etc. Note a component $k$ in this tree can be an internal node or a leaf, but every node has its associated probability $m_k$. A generating model for the tree is based on parameters $\boldsymbol{n}$ that govern the path taken from each node, and parameters $\boldsymbol{q}$ that govern when to stop in the tree, since an internal node also yields a valid index. The probability mapping is

$$m_k = q_k n_k \prod_{l \in \text{ancestors}(k)} n_l(1 - q_l)$$

where $q_k$ is the probability that one will remain at node $k$ and not descend to its children, and $n_l$ is the probability that child $l$ will be chosen. Note $q_k = 1$ for each leaf node $k$, and for each parent node $k$, $\sum_{l \in \text{children}(k)} n_l = 1$. The probabilities $q_k$ and $n_k$ form a dual representation for $m_k$ and the mapping is invertible.

To present the hierarchical version of discrete PCA then, we just need to give the sampling scheme for these $q_k$ and $n_k$. Then $\boldsymbol{m}$ is computed and the rest of the model proceeds as before. For each $k$ that is not a leaf, and its children given by $l_1, l_2, \ldots, l_{B_k}$ we have

$$q_k \sim \text{Beta}(\alpha_{1,k}, \alpha_{2,k}) ,$$
$$(n_{l_1}, n_{l_2}, \ldots, n_{l_{B_k}}) \sim \text{Dirichlet}(\boldsymbol{\beta}_k) .$$

By the well known compositional property of Dirichlets, if the parameters are set in a particular way (the sum of $\boldsymbol{\beta}_k$ equals $\alpha_{2,k}$), this flattens out into one big $K$ dimensional Dirichlet, which of course defeats the whole purpose! Thus we carefully set the parameters to avoid this. We use $\boldsymbol{\beta}_k = (1/B_k, 1/B_k, \ldots, 1/B_k)$ to induce the children to have quite varying proportions, and make $\alpha_{1,k}, \alpha_{2,k}$ something like $1, 10$ for the root node (a weak preference for 10% stop words) and $10, 60$ for lower parent nodes, a stronger preference for 14% topically shared words; the strength is essential to stop the hierarchy flattening and making the intermediate nodes topically coherent.

For a fixed tree structure, the Gibbs version of this scheme just modifies the sampling of $\boldsymbol{m}$, and thus is straightforward to implement. The mean field version now has dual parameters giving a Beta and Dirichlet approximation to the posteriors of $q_k$ and $n_k$ respectively. Using the mean field formulation of Ghahramani and Beal (Ghahramani & Beal, 2000) presented for discrete PCA in (Buntine, 2002), the development is tedious but straightforward, and has the same form as its non-hierarchical version.

### 3.5 INFERENCE ON NEW DATA

A typical use of the model requires performing inference related to a particular document. In our experience, the posterior Dirichlet approximations for $m$ produced by mean field can be extreme, and not good for inference, also discussed in (Minka & Lafferty, 2002). Minka *et al.* recommend using an approximation due to Cowell, Dawid and Sebastiani (1996) for this task. This does sampling to estimate both the means and variances of the proportions $m$ for a document, and then sets a Dirichlet to agree in mean, and to agree in average standard deviation. This approach can thus be used for data summarization of samples. Minka *et al.* also recommend importance sampling be used to perform inference about quantities related to a document, using an approximate Dirichlet as the importance distribution. For larger component counts $K > 20$, we have found this method fails: importance sampling is notoriously poor in large dimensions when the sampling distribution is not close to optimal.

We avoid this problem by using a minimum variance importance sampler. The trick described in Section 3.2 can be used here as well. Suppose, for instance, we wished to estimate how well a snippet of text, a query, matches a document. Our document's topics are summarised by the hidden variables $m$. If the new query is represented by $x$, then $p(x|m, \Omega, K)$ is the matching quantity we would like ideally. Fixing the model parameters $\Omega$ for now, our best estimate to this matching quality measure is given by the posterior expectation

$$\text{Exp}_{m \sim p(m|r, x, \Omega, K)} \left( p(x|m, \Omega, K) \right)$$

This can be estimated in few samples by drawing Gibbs samples $m_n$ of the hidden proportions from the modified posterior $p(m|r, x, \Omega, K)$ (note the query $x$ now occurs on the right hand side here as well as the original document data $r$) and computing

$$\frac{N}{\sum_n 1/p(x|m_n, \Omega, K)} \ .$$

As before, this sampling yields minimum variance estimates for any importance sampler.

Note using this method, we have observed that the approximate Dirichlet posteriors estimated by the mean field method can have a posterior variance over 100 times smaller than the real posterior variance–in use we have observed this translates to very poor performance.

## 4 EXPERIMENT WITH A HIERARCHY

An analysis was made of 34,449 documents crawled from .GOV containing the word "Iraq". About 50 common stop-words were removed, and only the top 76,525 words were retained, yielding 10,738,635 words in total. A hierarchy was built with branching factor 7 and depth 3. Gibbs using a burn-in of 100 cycles and final recording (of $\Omega$) of 100 cycles takes 4 hours and created 57 components. This is four times an equivalent flat computation using mean field. Note we might have settled on about

When inspecting the model, the word multinomials for components below a node can be averaged to give a representation of the $k$-th node

$$\frac{m_k}{m_k + \sum_{l \in \text{children}(k)} m_l} \Omega_{k, \cdot} +$$
$$\sum_{l \in \text{children}(k)} \frac{m_l}{m_k + \sum_{l \in \text{children}(k)} m_l} \Omega_{l, \cdot}$$

The resultant word probabilities were inspected by viewing the dominant words. Some of the components are described in Table 1. Nodes 1-7 lie directly below the root node, and some additional children are given as well. Inspection shows about 70% coherency between an intermediate node and its children. Note that while general coherency to the hierarchy was obtained, node 1 mixes two separate concepts.

Table 1: Sample Nodes in a Hierarchy

| NODE | DESCRIPTION |
|------|-------------|
| root | General frequencies of words in collection |
| node 1 | Accounting and Finance + Imagery |
| node 2 | Opinions, Presidential and otherwise on international issues |
| node 3 | US government issues |
| node 4 | Department of State briefings and reports |
| node 5 | Issues of concern to voters |
| node 6 | Congressional and Embassy home pages |
| node 7 | Government funding and funded programs. |
| child of 1 | NASA,NOAA,USGS, etc. |
| child of 1 | Funding, loan, grant issues in Government supported business |
| child of 2 | Voice of America (radio station) |
| child of 2 | Asian-Pacific issues |
| child of 6 | Bibliographic data |
| child of 7 | Funding and resources regulations (i.e., w.r.t "reconstruction") |

## 5 EXPERIMENTS ON INDEPENDENCE OF COMPONENTS

To explore ICA comparison described in Section 2.2, we applied discrete PCA in both its flat and hierarchical version. We used the original 20 Newsgroups data[2]. We performed Porter stemming and eliminated numbers (except years), and basic stop words (propositions, conjunctions, common verbs like "has"), and then eliminated words with less than 4 occurrences or in less than 3 documents. The result was 19,997 documents with 29,998 lexemes and 2,262,161 lexemes in total.

Standard Gibbs was performed with $K = 150$, and hierarchical Gibbs with maximum $K = 200$, a branching factor of 7 and a starting hierarchy of $K = 57$. The hierarchy has approximately the same number of leaves as the flat version. Experiments with $K = 50, 100, 150, 200, 300$ had previously determined $K = 150$ to be the best number of components for the flat model using the methods of Section 3.3. The Dirichlet prior on the $J$-dimensional component multinomials ($\Omega_{k,\cdot}$) was an empirical prior with means estimated via Laplace smoothing and a prior count of 1. Words in a component are approximately Zipf-like apriori. The Dirichlet prior on the $K$ component proportions $\boldsymbol{m}$ was a uniform Dirichlet with the prior count of 1, as was the prior in the hierarchical version for branch selection. For the $K = 200$ hierarchical version, the final layer of leaf nodes extended the most populous nodes in the 57-node starting hierarchy at cycle 50. 200 cycles was done for burn-in, and a following 100 cycles for recording. The 200 component hierarchical version took 280 minutes of a 1.5GHz AMD CPU with 500Mb of main memory under Linux, and the 150 component flat version took 202 minutes. By comparison, mean field on the 150 component flat model takes 35 minutes with a 100 cycles.

To evaluate the independence of components, we compute the pairwise correlation co-efficient between the discrete ICA components, which correspond to the mean of the $\boldsymbol{m}$ vector times the word count for the document $L$. This gives 11175 correlations for the flat version and 19900 correlations for the hierarchical version. The resulting correlation coefficients are given in Figure 1. "D" labels correlations for the flat $K = 150$ model, "T-T" labels correlations between internal nodes of the $K = 200$ hierarchical model, "B-B" between leaf nodes, and "T-B" between leaf and internal nodes. These correlation coefficients are mostly surprisingly small, indicating to some degree independence of components has been achieved. An inves-tigation into the interactions here using methods of (Jakulin & Bratko, 2003) revealed that in the hierarchical model much of the interaction is up and down the hierarchy. Children and parents can have some dependence.

[2]See http://www.ai.mit.edu/~jrennie/20Newsgroups/.

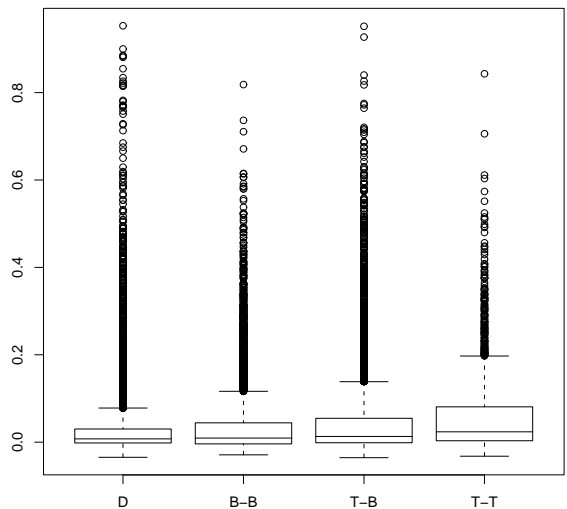

Figure 1: Box Plots for the Two Models

## 6 CLASSIFICATION EXPERIMENTS

We tested the use of discrete PCA in its role as a feature construction tool, a common use for PCA and ICA, and as a classification tool. For this, we used the 20 newsgroups collection described previously as well as the Reuters-21578 collection[3]. We employed the $SVM^{light}$ V5.0 (Joachims, 1999) classifier with default settings. For classification, we added the class as a distinct multinomial (cf. Section 3.1) for the training data and left it empty for the test data, and then predicted the class value using the method of Section 3.5. Note that for performance and accuracy, SVM is the clear winner: as the state of the art optimized discrimination-based system this is to be expected. It is interesting to see how discrete PCA compares.

Each component can be seen as generating a number of words in each document. This number of component-generated words plays the same role in classification as does the number of lexemes in the document in ordinary classification. In both cases, we employed the TF-IDF transformed word and component-generated word counts as feature values. Since SVM works with sparse data matrices, we assumed that a component is not present in a document if the number of words that would a component have generated is less than 0.01. The components alone do not yield a classification performance that would be competitive with SVM, as the label has no distinguished role in the fitting. However, we may add these component-words in the default bag of words, hoping that the conjunctions of words inherent to each component will help improve the classification performance.

[3]The Reuters-21578, Distribution 1.0 test collection is available from David D. Lewis' professional home page, currently: http://www.research.att.com/~lewis

For the Reuters collection, we used the ModApte split. For each of the 6 most frequent categories, we performed binary classification. Further results are disclosed in Table 2. No major change was observed by adding 50 components to the original set of words. By performing classification on components alone, the results were inferior, even with a large number of components. In fact, with 300 components, the results were worse than with 200 components, probably because of overfitting. Therefore, regardless of the number of components, the SVM performance with words cannot be reproduced by component-generated words in this collection.

Table 2: SVM Classification Results[4]

| CAT | SVM | | SVM+dPCA | |
|---|---|---|---|---|
| | ACC. | P/R | ACC. | P/R |
| earn | 98.58 | 98.5/97.1 | 98.45 | 98.2/97.1 |
| acq | 95.54 | 97.2/81.9 | 95.60 | 97.2/82.2 |
| moneyfx | 96.79 | 79.2/55.3 | 96.73 | 77.5/55.9 |
| grain | 98.94 | 94.5/81.2 | 98.70 | 95.7/74.5 |
| crude | 97.91 | 89.0/72.5 | 97.82 | 88.7/70.9 |
| trade | 98.24 | 79.2/68.1 | 98.36 | 81.0/69.8 |

| CAT | dPCA (50 comp.) | | dPCA (200 comp.) | |
|---|---|---|---|---|
| | ACC. | P/R | ACC. | P/R |
| earn | 96.94 | 96.1/94.6 | 97.06 | 96.3/94.8 |
| acq | 92.63 | 93.6/71.1 | 92.33 | 95.3/68.2 |
| moneyfx | 95.48 | 67.0/33.0 | 96.61 | 76.0/54.7 |
| grain | 96.21 | 67.1/31.5 | 97.18 | 77.5/53.0 |
| crude | 96.57 | 81.1/52.4 | 96.79 | 86.1/52.4 |
| trade | 97.82 | 81.4/49.1 | 97.91 | 78.3/56.0 |

Classifying newsgroup articles into 20 categories proved more successful. We employed two replications of 5-fold cross validation, and we achieved the classification accuracy of 90.7% with 50 additional dPCA components, and 87.1% with SVM alone. Comparing the two confusion matrices, the most frequent mistakes caused by SVM+dPCA beyond those of SVM alone were predicting talk.politics.misc as sci.crypt (26 errors) and talk.religion.misc predicted as sci.electron (25 errors). On the other hand, the components helped better identify alt.atheism

and talk.politics.misc, which were misclassified as talk.religion.misc (259 fewer errors) earlier. Also, talk.politics.misc and talk.religion.misc were not misclassified as talk.politics.gun (98 fewer errors). These 50 components were not very successful alone, resulting in 18.5% classification accuracy. By increasing the number of components to 100 and 300, the classification accuracy gradually increases to 25.0% and 34.3%. Therefore, many components are needed for general-purpose classification.

From these experiments, we can conclude that components may help with tightly coupled categories that require conjunctions of words (20 newsgroups), but not with the keyword-identifiable categories (Reuters). Judging from the ideas in (Jakulin & Bratko, 2003), the components help in two cases: a) when the co-appearance of two words is more informative than sum of informativeness of individual appearance of either word, and b) when the appearance of one word implies the appearance of another word, which does not always appear in the document.

# 7  INFERENCE EXPERIMENTS

We performed document retrieval using the new Reuters Corpus[5], containing over 806,791 news items from 1996 and 1997. Documents were preprocessed by a simple parser so that document data was presented to discrete PCA as 4 separate bags, one for nouns, one for verbs, one for adjectives and one for adverbs. These resulted in 155,325 distinct lexemes, but in four separate multinomials for a total of 92,639,516 lexemes in the full collection. A hierarchical model was built as before with a branching factor of 10 and 111 nodes in total, in an overnight run (whereas, a 1111-node hierarchy took 6 days on a dual CPU).

Queries were performed on the $K = 111$ model using the method of Section 3.5 applied by re-ranking the top 20,000 documents returned by TF-IDF. The inference for this typically take 5 minutes time in total for each query, and thus is not a realistic information retrieval method at present. We evaluated the results by inspection, and compared the results with state of the art TF-IDF processing from the Lemur Toolkit applied to the same preprocessed document data. Two queries are demonstrated here, the first extracted from an article about the 2003 UK housing market.

Query 1: *A shortage of properties across the country is adding pressure to the housing market, and keeping prices on the rise, a new report says.*

Query 2: *football fans in Germany.*

Table 3: Results for Query 1

Discrete PCA component matching

| | |
|---|---|
| UK | Property shortage squeezes UK housing market |
| UK | UK housing market strongest for eight years |
| UK | Housing shortage squeezes prices through roof |
| UK | UK housing market recovery continues, survey |
| UK | UK housing shortage drives prices higher |

TF-IDF Matching as per Lemur Toolkit

| | |
|---|---|
| USA | TEXT - Excerpts of the Feb FOMC meeting min. |
| UK | Full text of April 10 UK monetary minutes |
| SA | Text of President Mandela's speech to parliament |
| UK | Full text of October 30 UK monetary minutes |
| USA | Fed says economy growing steadily, prices muted |

Table 4: Results for Query 2

Discrete PCA component matching

| | |
|---|---|
| GDR | Soccer-German wonder-winger Libuda dies at 52 |
| GDR | Soccer showcase-contrasting priorities for two Borussias |
| GDR | Soccer-fans pray for Borussia triumph |
| GDR | Soccer-Schalke fans party all night |
| GDR | Soccer-Sammer tipped to win footballer of the year |

TF-IDF Matching as per Lemur Toolkit

| | |
|---|---|
| UK | Yearend-Football home, but Estonia missing |
| UK | Yearend-Football home, but Estonia missing |
| GDR | Soccer-Dortmund, Schalke bring cheer to troubled Ruhr |
| GDR | Soccer-Eintracht Frankfurt battling to avoid obscurity |
| CZ | Soccer-Czech fans harken back to old days as pay-TV |

In Tables 3 and 4, five discrete PCA results are first, then five TF-IDF results. This inference method is in the spirit of language modelling for information retrieval (Croft & Lafferty, 2003) and it is clear that topical content is being retrieved. Note that articles retrieved using the mean field approximation to perform inference were poor, as expected from Section 3.5.

Generally some results were impressive and some patchy but not much worse. We speculate that in queries of a more general nature, better results seem to be returned, but in others TF-IDF is clearly preferable due to its speed. Statistical inference must fail in specific queries where keywords are insufficiently dense to allow a statistical trace to be recovered.

## 8   CONCLUSION

We have argued that discrete PCA is a discrete version of ICA. We have shown it performs as a data reduction tool, though somewhat slow, and that it can be used for inference as well, for instance for information re-trieval. A hierarchical version developed retains many of the same properties but develops components with a hierarchical coherence, better for human interpretation. This required some theoretical development, as discussed, including optimal sampling for choosing the right number of components, and optimal sampling for inference. The software demonstrated here is available under the GNU GPL license from the first author.

**Acknowledgements**

The first author's work was supported by the Academy of Finland under the PROSE Project, and by Finnish Technology Development Center (TEKES) under the Search-Ina-Box Project. The second author wishes to thank his advisor Ivan Bratko.

**References**

Blei, D. M., Ng, A. Y., & Jordan, M. I. (2003). Latent Dirichlet allocation. *Journal of Machine Learning Research*, *3*, 993–1022.

Buntine, W. (2002). Variational extensions to EM and multinomial PCA. *ECML 2002*.

Croft, W., & Lafferty, J. (Eds.). (2003). *Language modeling for information retrieval*. Kluwer Academic.

Ghahramani, Z., & Beal, M. (2000). Propagation algorithms for variational Bayesian learning. *NIPS* (pp. 507–513).

Hofmann, T. (1999). Probabilistic latent semantic indexing. *Research and Development in Information Retrieval* (pp. 50–57).

Hofmann, T., & Buhmann, J. M. (1997). Pairwise data clustering by deterministic annealing. *IEEE Transactions on Pattern Analysis and Machine Intelligence*, *19*, 1–14.

Hyvärinen, A., Karhunen, J., & Oja, E. (2001). *Independent component analysis*. John Wiley & Sons.

Jakulin, A., & Bratko, I. (2003). Analyzing attribute dependencies. *PKDD 2003* (pp. 229–240). Springer-Verlag.

Joachims, T. (1999). Making large-scale SVM learning practical. In B. Schölkopf, C. Burges and A. Smola (Eds.), *Advances in kernel methods - support vector learning*. MIT Press.

Minka, T., & Lafferty, J. (2002). Expectation-propagation for the generative aspect model. *UAI-2002*. Edmonton.

Pritchard, J., Stephens, M., & Donnelly, P. (2000). Inference of population structure using multilocus genotype data. *Genetics*, *155*, 945–959.

Woodbury, M., & Manton, K. (1982). A new procedure for analysis of medical classification. *Methods Inf Med*, *21*, 210–220.